\documentclass[conference]{IEEEtran}

\usepackage{graphicx, amsmath, multirow}

\begin{document}


\title{Optimisation of Air-Ground Swarm Teaming for Target Search, using Differential Evolution}

\author{
    \IEEEauthorblockN{Jiangjun Tang}
    \IEEEauthorblockA{School of Engineering and IT\\UNSW Canberra,Australia\\Email: j.tang@adfa.edu.au}
    \and
    \IEEEauthorblockN{George Leu}
    \IEEEauthorblockA{School of Engineering and IT\\UNSW Canberra, Australia\\Email: g.leu@adfa.edu.au}
    \and
    \IEEEauthorblockN{Yu-Bin Yang}
    \IEEEauthorblockA{State Key Laboratory for Novel Software Technology \\ Nanjing University, Nanjing, China\\Email: yangyubin@nju.edu.cn}
}


\maketitle

\makeatletter
\def\ps@IEEEtitlepagestyle{
  \def\@oddfoot{\mycopyrightnotice}
  \def\@evenfoot{}
}
\def\mycopyrightnotice{
  {\footnotesize
  \begin{minipage}{\textwidth}
  \centering
  Copyright~\copyright~2019 IEEE. Personal use of this material is permitted.  Permission from IEEE must be obtained for all other uses, in any current or future media, including reprinting/republishing this material for advertising or promotional purposes, creating new collective works, for resale or redistribution to servers or lists, or reuse of any copyrighted component of this work in other works.
  \end{minipage}
  }
}

\begin{abstract}
This paper presents a swarm teaming perspective that enhances the scope of classic investigations on survivable networks. A target searching generic context is considered as test-bed, in which a swarm of ground agents and a swarm of UAVs cooperate so that the ground agents reach as many targets as possible in the field while also remaining connected as much as possible at all times. To optimise the system against both these objectives in the same time, we use an evolutionary computation approach in the form of a differential evolution algorithm. Results are encouraging, showing a good evolution of the fitness function used as part of the differential evolution, and a good performance of the evolved dual-swarm system, which exhibits an optimal trade-off between target reaching and connectivity.
\end{abstract}



\IEEEpeerreviewmaketitle

\section{Introduction}

The use of mobile aerial agents to support ground communication has gained significant traction in recent years, especially due to advances in drone technology. Typically, current research in this direction considers  unmanned aerial vehicles (UAVs) that act as relays to facilitate wireless communication between mobile ground agents operating on the ground. The success of the UAV support, i.e. the ground network survivability, is given by the extent to which the ground agents remain connected in spite of various sources of disruption, thanks to the convenient positioning of the UAVs.

One of the main assumptions in studies on survivable networks is that the air and ground swarms operate as separate swarms, with their own purposes and mobility models. Thus, the UAV swarm only acts as a facilitator, which means that the ground swarm operates in the field to accomplish a certain task and is unaware of the UAVs' existence. As a consequence, the behaviour of the UAVs is influenced by the behaviour of the ground agents, but not the other way around. Also, the UAVs are unaware of the task the ground agents need to perform; they only offer communication support based on the perceived movement of the ground agents. Arguably, this is sufficient in a narrow view on survivable networks, where only ground communication is at stake. However, we believe that is pertinent and useful to attempt to integrate the two swarms as much as possible, and consider not only a one way facilitation, but true cooperation. That means, the UAV swarm could (and should, we argue) offer communication support while also explicitly participating in the accomplishment of the ground task, which involves a two way interaction between the swarms. We see this integration as a two fold endeavour: (1) as an attempt to model the two types of agents (i.e. the swarms) using similar, if not identical, conceptual representations and (2) model the two way interactions (i.e. cooperation) to optimise system performance beyond the narrow survivability view.

Thus, in this paper we promote a swarm teaming perspective that enhances the scope of classic investigations on survivable networks. We consider a target search context in which the ground swarm has to search for and reach as many targets as possible in the field. The UAVs provide connectivity support, which is an indirect way of facilitating the success of search operations, but also directly search for targets, acting as range extensions for the sensing capabilities of the ground agents. From an interaction point of view, each agent type influences and is influenced by the other type. From a goal achievement point of view two objectives are followed in the same time: maintaining ground connectivity (which is the typical survivable networks goal) and the number of targets reached. To optimise the system against both these objectives in the same time, we use an evolutionary computation approach in the form of a differential evolution (DE) algorithm. The generic ``target search'' context is pertinent to numerous real-world scenarios and applications, such as finding and treating survivors in disaster areas, finding and exploiting resources, and many others.

The rest of the paper is organized as follows. Section~\ref{Section:Background} briefly discusses the main achievements reported in the literature in relation to survivable networks. Then, Section~\ref{Section:Methodology} describes the methodology used to model the agents and optimise the system in relation to the two objectives. Further, Section~\ref{Section:Results} presents and discusses the experiment results, and, in the end, Section~\ref{Section:Conclusion} summarises the findings and concludes the paper.

\section{Background} \label{Section:Background}

Traditionally, research on survivable networks concentrates on the mobility models adopted by the UAVs in an aerial swarm to facilitate communication between the agents of the ground swarm. These refer to the decision-making mechanisms that provide the optimal air trajectories and/or positions. As a historical note, the aerial support for ground communication first used satellites and high-altitude fixed-wing aircraft. However, these have substantial limitations from multiple points of view, such as mobility or scalability, and thus, the concept of swarms of aerial agents can be hardly considered.
The use of UAV swarms was only possible when substantial progress in drone technology have been achieved~\cite{Leu2019icsi,Leu2019cec}. Availability of miniaturised, versatile and low-cost multi-copter drones made possible the investigation of systems with larger numbers of UAVs, and also opened doors for using swarm intelligence mobility models for these UAVs. The increased number of agents also raised the question about the information available to UAVs, where information can be global, as proposed in~\cite{Kuiper2006,Elston2008}, or local as proposed in~\cite{BasuUAV2004,Hauert2009}, where inputs come only from the neighboring agents. However, most of the literature concentrates on models that use local information, which will be discussed further. 

The most relevant studies using local information can be grouped in three major categories: random models, parametric/mathematical models and nature-inspired models. The random models were initially based on pure random trajectories, as discussed in~\cite{Kuiper2006}. More recent work in this direction proposed the so-called chaos-enhanced mobility, which build on the early random approaches. Chaos-enhanced mobility generated a considerable amount of research in the recent literature~\cite{Rosalie2016,Rosalie2017,Rosalie2018}. The parametric/mathematical approaches consider analytic methods with fixed pre-tuned parameters. Many of these studies date from the early times of survivable network research~\cite{KarInria2003,BasuUAV2004,Hui2011,Cetin2012,Goddemeier2012}, but few recent ones exist too~\cite{ZhouAGCop2015, Zhang2018}. The nature-inspired methods consider swarming behaviours, such as ant colony pheromone-based mobility~\cite{Kuiper2006}, bat algorithms~\cite{SUAREZ2018}, or boids-based finite state machines~\cite{BasuUAV2004}. Artificial evolution have been also used, with evolutionary computation techniques evolving parameters of the controllers used in aerial agents~\cite{Hauert2009}. This direction has been less investigated because of the slow convergence of evolutionary computation techniques~\cite{Tang2012}, especially when large numbers of parameters need to be optimised. Instead of the sole use of evolutionary computation, hybrid methods have been proposed recently, which use boids-based flocking behaviours (i.e. based on Reynolds' boids~\cite{Reynolds1987}) to implement UAVs' movement and evolutionary computation with reduced number of parameters to optimise their movement towards supporting ground communication. Such hybrid methods have been used very recently~\cite{Leu2019cec,Leu2019icsi}, and proved to be usable in real-time contexts.


Most of the existing studies mentioned above consider the UAVs and ground agents as separate groups, and as a result, their operation is implemented using different conceptual models. Ideally, as explained in the introductory session, it would be desired that aerial and ground agents operate based on identical, or at least similar mobility models. This integration is beneficial especially from a swarm teaming perspective, where the different types of agents are aware of each-other's tasks and can cooperate to achieve those tasks together. Thus, it would be desirable that the swarms share certain behavioural features, such as their interaction architecture (i.e. boids-based neighborhoods and forces). In the survivable networks literature there is very little preoccupation for this aspect. Only very recently, in~\cite{Leu2019cec}~and~\cite{Leu2019icsi} the authors take some steps towards a certain degree of integration. In these studies the UAV and ground swarms are both modelled using boids-based swarms which share the key aspects of flocking-boid agents: the neighborhood and the interaction forces. The authors claim that the two swarms can be treated as one dual-level swarm, with two sub-swarms having their own settings but common operational (i.e. mobility) model. While this is a step forward from the typical approach in survivable networks, it still does not account for a true two-way interaction between the swarms. The UAV swarm is still a facilitator only, and the influence is one one-way (i.e. only the UAVs are influenced by the ground agents' movement).

In light of the above discussion, in this paper we take another step, and enhance the concepts presented in~\cite{Leu2019cec,Leu2019icsi} with a two way interaction between swarms, which makes the conceptual model of operation appropriate for real-world contexts like like search and rescue, resource seeking and exploitation, military operations, etc.

\section{Methodology}\label{Section:Methodology}

\subsection{Environment and goals}
The environment we use in this paper is typical for research on survivable networks, having been used in various studies over time~\cite{BasuUAV2004,Gaudiano2005,Leu2019cec,Leu2019icsi}. This is a $2D$ rectangular simulation space $S$ (i.e. a plane surface) of length $L$ and width $W$, which is populated with obstacles and targets for the ground agents, as explained in detail later in the experiment design section. The ground agents operate in the simulation space avoiding collisions with each-other, avoiding obstacles and seeking targets. The UAVs move over the simulation space with no restriction, since no obstacles exist in the air, except avoiding collisions with each-other; they also seek for targets. The origin of the space, which is used for force, velocity, and position vector calculation, is the bottom left corner.

The behaviours of the two types of agents, UAVs and ground, are modeled based on the classic boids model of Reynolds~\cite{Reynolds1987}, which means all agents are flocking boids. We use the two key concepts of the original boids model, i.e. the neighborhood-based interaction and the three boids forces: cohesion, alignment, and separation. However, the concepts are altered differently for each of the two swarms, to suit the context investigated in this paper. The resultant agent features are discussed below, in Sections~\ref{Subsection:GroundRules}~and~\ref{Subsection:AirRules}).

As explained in the introductory section, we consider two objectives that the swarms need to achieve together: communication between ground agents is maintained as much as possible at all times, and the highest possible number of targets is reached.

The communication between ground agents is evaluated based on the ``\textit{connectivity}'' concept. This uses the graph theoretical concept of connected network component~\cite{Albert2002}, which denotes a sub-graph where any two nodes are connected to each other. Connectivity is the number of connected components that exist at a moment in time within the swarm of ground agents. Ideally, when the ground swarm is fully connected, the connectivity has the value $1$, which means that there is only one sub-graph equal to the entire ground network. In practice, more than one connected component may exist at a certain moment in time, but among them one or several giant connected components should exist~\cite{Newman2002}. In this paper, we measure the survivability by calculating the size of the largest (giant) connected component throughout the simulation.

The targets appear in the field one after the other at random positions, so that only one of them is present in the field at moment in time. A target is considered ``reached'' when a number $n_r$ of ground agents arrive at and touch the target. When a target is reached, it disappears and a another one is generated. This describes contexts like search and rescue, where a survivor is needs to be reached by several rescuers to be properly treated, or resource seeking, where a source of goods is found and several exploiting agents are needed to harvest it until depletion.

\subsection{UAVs} \label{Subsection:AirRules}

The swarm of $m_a$ airborne agents is denoted by $A=\{A_i|\forall i=1..m_a\}$. This is a boid-based swarm, where the behavior of each individual UAV is governed by a network-based neighbourhood~\cite{Tang2018}, and six forces.

\subsubsection{Neighbourhood}
The neighbourhood used for UAV force computation is calculated based on a communication range $R$, which means an UAV senses the presence of other agents (UAVs or ground) up to a distance equal to the communication range $R$, over a 360 degree angle (a detailed discussion on network-based versus classic vision-based boids can be found in~\cite{Tang2018}). The sensing capabilities also include a target detection range $R_{AT}$.

\subsubsection{Forces}
The first three forces are air-to-air cohesion, alignment, and separation, representing the influence received from the peer UAVs. These forces are denoted as $C_{AA}$, $A_{AA}$ and $S_{AA}$ respectively. The separation distance associated to the air-to-air separation force is denoted $SD_A$. The next two forces are ground-to-air cohesion and alignment, representing the influence received from the ground agents situated in the neighbourhood. These are denoted as $C_{GA}$ and $A_{GA}$ (a ground-to-air separation force is not deeded because there is no possibility for a collision between UAVs and ground agents). The last force is attraction to a target situated in the neighbourhood. This is denoted with $T_A$ and is similar to a cohesion force; it applies when the target is in the target detection range $R_{AT}$, but the UAV is attracted only by the target instead of a group of neighbors. As a result, the direction of $T_A$ is towards the target center instead of the center of mass of neighbouring entities. All six forces are calculated like in the original study of Reynolds~\cite{Reynolds1987}.

In addition to the communication with ground agents, the airborne agents can also communicate directly with other airborne agents, which means the UAV swarm operates as an ad-hoc network.

\subsubsection{Velocity and position update}
Updating the velocity of an UAV employs a weighted sum of all forces applied to it. A weighted sum with six terms is considered, corresponding to the six forces mentioned above. In this sum all forces are normalised. Thus, the velocity ($V_{A_i}$) of an UAV ($A_i$) can be expressed as in Equation~\ref{Equation:AirVelocity}:

\begin{equation}
    \begin{split}
        V_{A_i}(t) = & V_{A_i}(t-1) + \\
            & + W_{C_{AA}} C_{{AA}_i}(t) + W_{A_{AA}} A_{{AA}_i}(t) + W_{S_{AA}} S_{{AA}_i}(t)\\
            & + W_{C_{GA}} C_{{GA}_i}(t) + W_{A_{GA}} A_{{GA}_i}(t) + \\
            & + W_{T_A} T_{A_i}(t)
    \end{split}
	\label{Equation:AirVelocity}
\end{equation}
where $Cs$, $As$, $S$, and $T$ are the normalised force vectors, and $Ws$ are the weights of the force vectors.

Then, the position $P_{A_i}$ at time $t$ of each airborne agent $A_i$ can be updated as in Equation~\ref{Equation:AirPosition}, where the position vector has always its origin in the simulation space origin, and the velocity vector at time $t$ has its origin in the position vector tip at time $t-1$ (i.e. a tip-to-tail vector addition is performed). If the new position of an agent (position vector tip) is outside the boundary of the space $S$, a reflection rule is applied to keep the agent within the simulation space; this is a simple reflection, such as a ray of light in the mirror, or a ball that hits a wall.

\begin{equation}
	P_{A_i}(t) = P_{A_i}(t-1) + V_{A_i}(t)
	\label{Equation:AirPosition}
\end{equation}

The rules considered above for the airborne agents allow them to move according to a swarming behavior, where they are influenced by the neighboring agents' status. However, just modeling the airborne agents as a swarm does not lead to optimality. In order to do that an optimisation of the weights is needed. The optimisation algorithm is described in detail in Section~\ref{Section:Optimization}.

\subsection{Ground Agents} \label{Subsection:GroundRules}

The swarm of $n_g$ ground agents is denoted by $G=\{G_i|\forall i=1...n_g\}$. This is a boid-based swarm, where the behavior of each individual agent is governed by a classic vision-based neighbourhood~\cite{Reynolds1987}, and seven forces.

\subsubsection{Neighbourhood}
The neighbourhood used for force computation is calculated based on vision, which includes a visual distance $V_d$ and a visual angle $V_\alpha$. The sensing capabilities also include an obstacle detection range $R_O$ and a target detection range $R_{GT}$, respectively.

\subsubsection{Forces}

The first three forces are ground-to-ground cohesion, alignment and separation, representing the influence received from the peer ground agents. These forces are denoted as $C_{GG}$, $A_{GG}$ and $S_{GG}$ respectively. The separation distance associated to the ground-to-ground separation force is $SD_G$. The next two forces are air-to-ground cohesion and alignment, representing the influence received from the UAVs that the ground agent is connected to directly. These are denoted as $C_{AG}$ and $A_{AG}$ (an air-to-ground separation force is not deeded because there is no possibility for a collision between UAVs and ground agents). The next force is attraction to a target situated in the neighbourhood. This is denoted with $T_G$ and is similar to the one of the UAVs except the target detection range is $R_{GT}$. All six forces are defined like in the original study of Reynolds~\cite{Reynolds1987}.

\begin{figure}[h]
    \centering
    \includegraphics[width=0.7\linewidth]{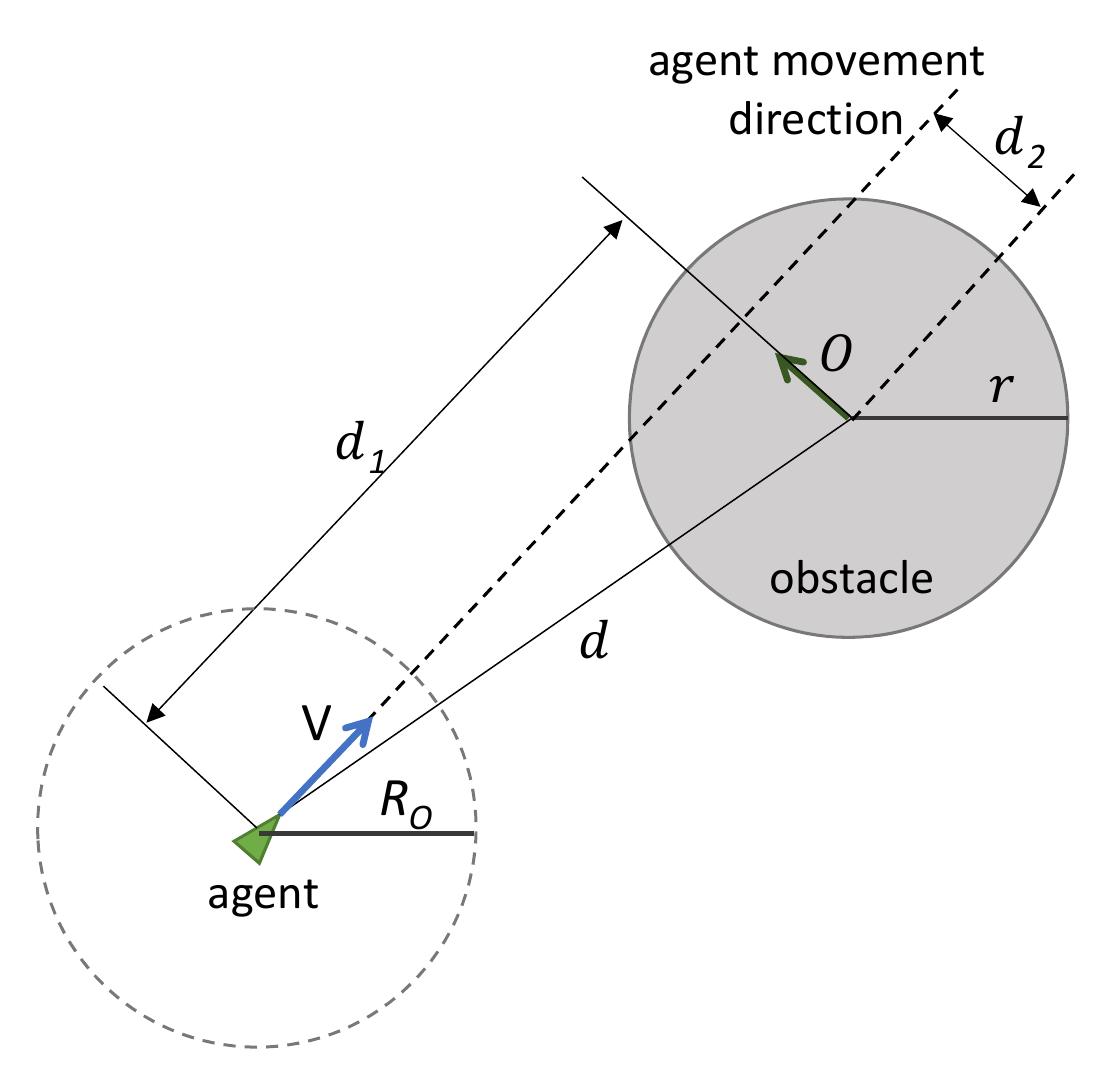}
    \caption{Obstacle detection and avoidance of a ground agent}
    \label{fig:obstacleavoidance}
\end{figure}

The seventh force is the ground obstacle avoidance force ($O_G$). This ensures the agent steers away from an obstacle in the environment. The obstacles and the corresponding force are modelled based on the method described in~\cite{Reynolds1999steering}. As stated in~\cite{Reynolds1999steering}, an obstacle of any shape can be approximated roughly by its bounding circle, or in more detail by a convenient aggregation of multiple circles. Thus, in this paper we consider for convenience simple obstacles of a circular shape.
Figure~\ref{fig:obstacleavoidance} depicts the obstacle detection and avoidance scheme. As shown in the figure, aground agent is moving with a velocity $V$, and can detect obstacles in a range $R_O$. An obstacle is a static disc of radius $r$. $O_G$ is perpendicular on agent's movement direction. The force is applied only if the obstacle is in agent's detection range ($d_1<R_O$) and intersects its direction of movement ($d_2<r$).

\subsubsection{Velocity and Position Update}
Similar to the UAV case, the velocity $V_{G_i}$ of a ground agent $G_i$ at time $t$ is updated according to Equation~\ref{Equation:GroundVelocity}.

\begin{equation}
 \begin{split}
        V_{G_i}(t) = & V_{G_i}(t-1) + \\
            & + W_{C_{GG}} C_{{GG}_i}(t) + W_{A_{GG}} A_{{GG}_i}(t) + W_{S_{GG}} S_{{GG}_i}(t)\\
            & + W_{C_{AG}} C_{{AG}_i}(t) + W_{A_{AG}} A_{{AG}_i}(t) + \\
            & + W_{T_G} T_{G_i}(t) + \\
            & + W_{O_G} O_{G_i}(t)
    \end{split}
	\label{Equation:GroundVelocity}
\end{equation}
where $Cs$, $As$, $S$, $O_G$, $T_G$ are the normalised force vectors, and $Ws$ are the weights of the force vectors.

Based on the velocity calculated in Equation~\ref{Equation:GroundVelocity}, the position $P_{G_i}$ at time $t$ of each ground agent $G_i$ can be updated as in Equation~\ref{Equation:GroundPosition}, using the same vector addition method like in the UAV case. The ground agents are also bounded to the operation space through a reflection rule, which is similar to that applied to UAVs.

\begin{equation}
    P_{G_i}(t) = P_{G_i}(t-1) + V_{G_i}(t)
    \label{Equation:GroundPosition}
\end{equation}

Unlike the UAV case, for the ground agents not all weights will need to be optimised. We consider that only the two forces representing the influence from UAVs need to be optimised. All other forces describe a baseline searching behaviour which manifests regardless of the UAVs' existence. Therefore, the corresponding weights are fixed throughout the simulation; the values are presented in Table~\ref{tab:groundweights}.
\begin{table}[h]
    \caption{The ground movement patterns,implemented via various force weights applied to ground agents.}
    \centering
    \begin{tabular}{|l|r|} \hline
        Force weights & value \\ \hline
        Cohesion & 0.01 \\ \hline
        Alignment & 0.125 \\ \hline
        Separation & 1 \\ \hline 
        Obstacle avoidance & 1 \\ \hline
        Target Tracking & 1 \\ \hline 
    \end{tabular}
    \label{tab:groundweights}
\end{table}

\section{Optimisation and cooperation}

\subsection{Cooperation}
As explained in the introductory section, the two swarms need to achieve two objectives through cooperation: ground network survivability and number of targets reached. Maintaining the ground communication is achieved by the UAVs by following the movement of ground agents ad conveniently positioning to maximise connectivity. This is done by considering the two ground-to-air forces that influence the behaviour of the UAVs. The target reaching is achieved by the ground agents (1) through own search capabilities represented by their ground target attraction force, and (2) by following the movement of the UAVs, which can search targets too through their air target attraction force. The influence received from the UAVs is captured through the two air-to-ground forces of the ground agent model~\footnote{When we say influence we refer to the information received by ground agents from an UAV about the position of the target, via communication}.

We note that our approach does not contain an explicit cooperation mechanism. Instead, an implicit cooperation effect emerges due to the forces that create mutual influences between agents of the two swarms. Thus, the optimisation process, which we describe in the following sections, optimises the weights of those forces involved in this implicit mutual influencing process.

\subsection{Optimisation via differential evolution}\label{Section:Optimization}

To optimise the behaviours of the two swarms towards the two objectives, we use differential evolution in an offline setting, i.e. for each solution generated by the evolutionary algorithm the whole simulation is run. That means, the optimal solution found by the DE algorithm is a set of optimal parameters of the individual agents, which lead to the emergence of a collective cooperative behaviour that reaches a high number of targets while maintaining high connectivity. The parameter values are the same for all agents of the respective swarms.

\subsubsection{Chromosome}
A candidate solution in the DE algorithm is defined by (1) the weights of all UAV forces except air-to-air separation, (2) the air-to-air safe distance $SD_A$, (3) the speed of the UAVs ($s_a$), and (3) the two air-to-ground forces of ground agents. Thus, the chromosome is a 9-component tuple, as shown in Equation~\ref{Eq:Chromosome}.

\begin{equation}
	\begin{split}
		ch = ch\{ & W_{C_{AA}}, W_{A_{AA}}, SD_A, W_{C_{GA}}, W_{A_{GA}}, W_{T_A},s_a, \\
			  	  & W_{C_{AG}}, W_{A_{AG}} \}
	\end{split}
	\label{Eq:Chromosome}
\end{equation}

\subsubsection{Fitness}
The fitness function used by the DE is a weighted sum of the two objectives, and is modelled as below:
\begin{equation}
    f= W_n N_C + W_t \frac{N_T}{s_s}
    \label{Eq:Fitness}
\end{equation}
where $N_C$ is the average percentage of ground agents in the largest connected component during the simulation, and $N_T$ is the number of targets tracked by the ground agents. Since $N_C$ is always between 0 and 1, a scaling scalar value ($s_s=10$) is used for the second term of the sum to bring it into the same variation interval.

\subsubsection{Differential evolution with elitism}
A population of individuals (candidate solutions) is initially generated. Each of the individuals in this population is represented by the chromosomes discussed earlier. Let us denote the chromosome as shown in Equation~\ref{Eq:chromsome}:

\begin{equation}
    X_{i,G} = \left[x_{1,i,G},x_{2,i,G},x_{3,i,G},\dots, x_{j,i,G} \right] \quad j=1,2,\dots,N
    \label{Eq:chromsome}
\end{equation}
where, $i$ is the individual number, $j$ is the parameter number, $G$ is the generation number and $N$ is the total number of parameters. Each parameter is initialized with a random number within the range $min_j \leq x_j,i,1 \leq max_j$, where $min_j$ and $max_j$ are predefined ranges for each parameter type.

At each generation, two best individuals were kept: one is the best individual for target tracking and another is the best for network connectivity. And then, for each individual in the population three different individuals, $x_{i,r1}, x_{i,r2}, x_{i,r3}$, are randomly selected from the population and are treated as vectors. The weighted difference of two of the selected vectors is summed with the remaining vector, as shown in Equation~\ref{Eq:Donor}, to produce a donor vector. In this equation $v_{i, G+1}$ is the donor vector corresponding to the individual $x_{i,G}$ and $F$ is the scaling factor, a number between 0 and 2.
\begin{equation}
    v_{i,G+1}=x_{i,r1}+F(x_{i,r2}-x_{i,r3})
    \label{Eq:Donor}
\end{equation}

Once the donor vector $v_{i,G+1}$ has been generated, a trial vector $u{i,G+1}$ is generated by combining elements from both the donor vector and the target vector $x_{i,G}$ based on Equation~\ref{Eq:Mutation}. Element $j$ in the trial vector is equal to the element $j$ of the donor vector if a random number between 0 and 1 is less than or equal to the crossover rate, $CR$. If the random number is greater than $CR$ then element $j$ in the trial vector is equal to element $j$ in the target vector.

\begin{equation}
    u_{j,i,G+1}= \begin{cases}
        v_{j,i,G+1} &  {\rm rand}([0,1])\ \leq\ {\rm CR} \cr
        x_{j,{\rm i},G} & {\rm rand}([0,1])\ >\ {\rm CR} 
    \end{cases}
\label{Eq:Mutation}
\end{equation}

Once the trial is evaluated by our simulation, its fitness can be calculated using Equation~\ref{Eq:Fitness} and compared against the fitness of the original individual $x_{j,{\rm i,G}}$. The one with the larger fitness value is selected for the next generation.

\subsection{Experimental settings}
The following settings are applicable to the elements discussed in the methodology.\\
\textbf{Operational settings} The operation space is a $1000 \times 1000$ units square, where $n_g=100$ ground agents and $n_a=4$ airborne agents operate for 10000 simulation time-steps.\\
\textbf{Ground agents:} The ground agents have vision range $V_d=30$ units, vision angle $V_\alpha=360$ degrees, omnidirectional obstacle detection range of radius $R_O=30$ units, and omnidirectional target detection range of radius $R_T=30$ units. The safety distance used by the separation force is $SD_G=10$ units for all agents. The speed of ground agents is constant $|V_g|=1 unit/timestep$ throughout the simulation for all agents. The agents are initialised at the beginning of each simulation with random positions and random direction of the velocity vectors.\\
\textbf{Airborne agents:} The UAVs have a communication range of $R=300$ units. The weight of the air-to-air separation force is fixed to $fW_{S_{AA}}$ to implement an implicit collision avoidance mechanism between UAVs. The UAVs are initialised at fixed positions around the centre of the operation space, to form a full connected network. Their velocity (speed and direction) is randomly initialised.\\
\textbf{Obstacles:} There are five obstacles in the field, of circular shape of radius $r=80$. The obstacles have fixed positions, as follows: one in the centre of the space, and four in the centres of each quadrant.\\ 
\textbf{Targets:} The targets are placed at random positions in the space, and are only active one at a time. A target remains active until is reached by at least $n_r=10$ ground agents, then is deactivated (disappears) and another target is placed (appears) in the space.\\
\textbf{DE settings:} The population size is 50 and number of generations is 100. The donor factor $F$ is 0.6 and the crossover rate $CR$ is 0.8. The value range for genes in the chromosomes are given in Table~\ref{tab:ChromValueRanges}, and are typical for boid-based swarming agents, as reported in previous studies~\cite{Tang2018,Leu2019cec}. \\
\textbf{Randomness:} All elements and processes that use random assignment of values follow the uniform distribution, to ensure a uniform spread of the values within their designated ranges. A total of 20 simulation runs are conducted, with different random generator seeds for the evolutionary algorithm, to ensure statistical validity of the results. The seeds for agent initialisation are kept constant to ensure traceability and consistency of results over various simulation runs.

\begin{table}[h]
    \centering
    \caption{Value range for each gene in a chromosome}
    \begin{tabular}{|l|l|r|r|} \hline
         & Gene & Min & Max  \\ \hline
         UGV & alignment to air    &   0 & 0.1 \\ \cline{2-4}
             & cohesion to air     &   0 & 0.1 \\\hline
         UAV & alignment to air    &   0 &   1 \\ \cline{2-4}
             & cohesion to air     &   0 &   1 \\ \cline{2-4}
             & separation distance & 100 & 290 \\ \cline{2-4}
             & speed               &   1 &   5 \\ \cline{2-4}
             & alignment to ground &   0 &   1 \\ \cline{2-4}
             & cohesion to ground  &   0 &   1 \\\cline{2-4}
             & attraction target   &   0 &   1 \\  \hline
    \end{tabular}
    \label{tab:ChromValueRanges}
\end{table}

 
\section{Results and Discussion}\label{Section:Results}

\subsection{Evaluation of the DE algorithm}
\begin{figure}
    \centering
    \includegraphics[width = 0.48\textwidth]{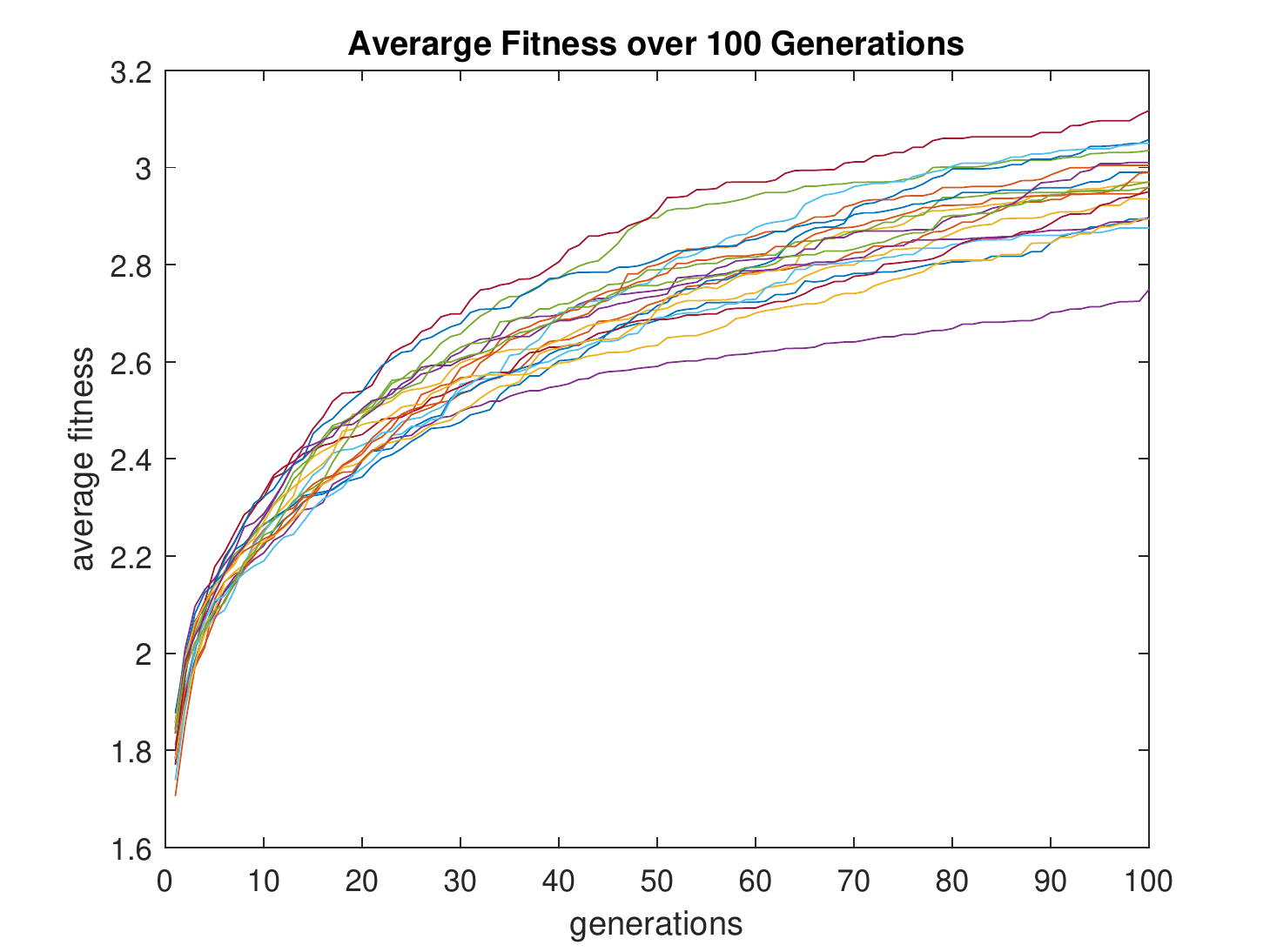}
    \caption{Average Fitness over 100 generations of each run}
    \label{fig:AvgFitnessAll}
\end{figure}

\begin{figure}
    \centering
    \includegraphics[width = .48\textwidth]{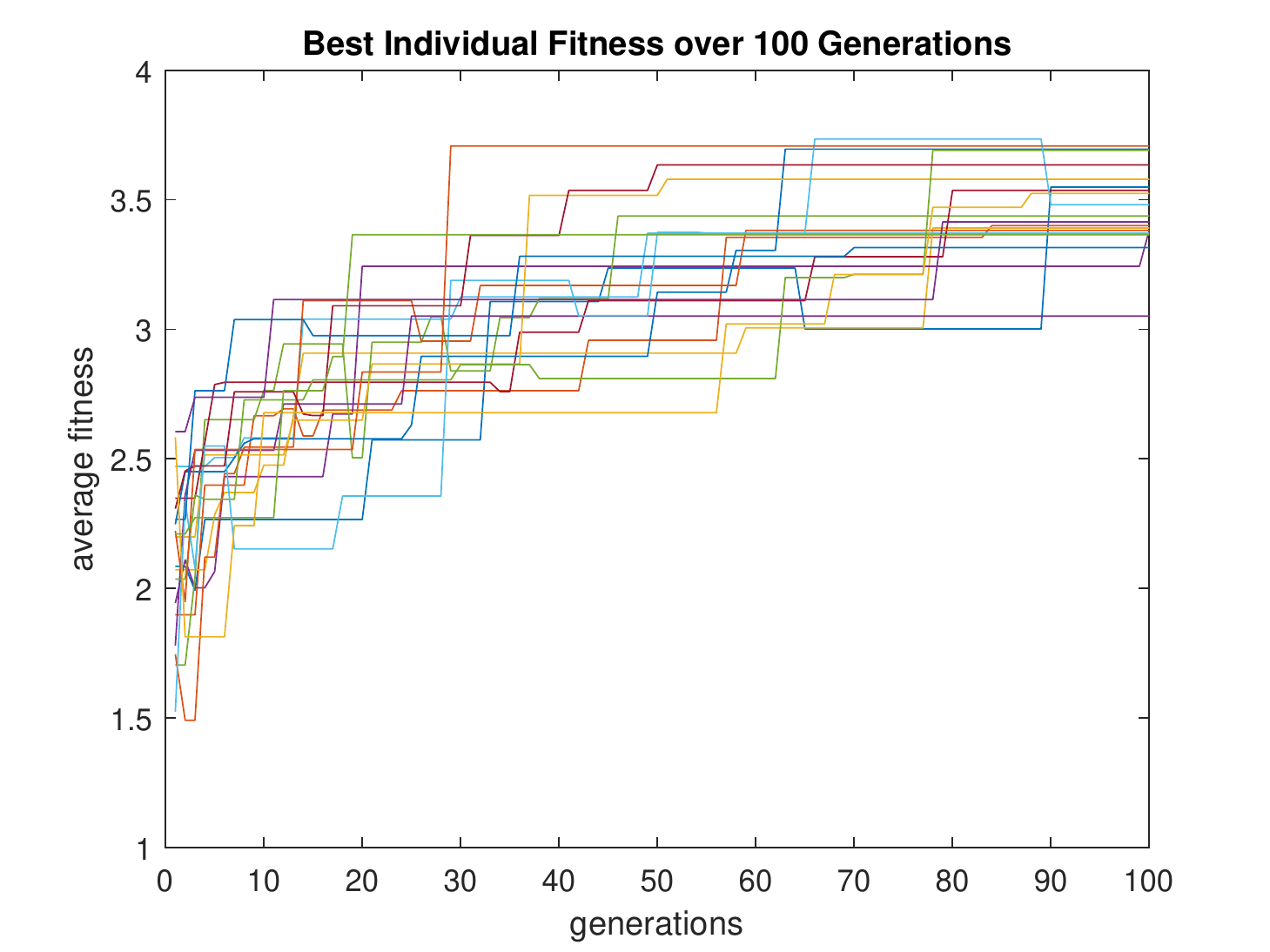}
    \caption{Best individual fitness over 100 generations of each run}
    \label{fig:BestFitness}
\end{figure}

To evaluate the proposed method we first collate the results showing how the optimisation process works. Figure~\ref{fig:AvgFitnessAll} shows for each of the 20 simulation runs how fitness improves over 100 generations of differential evolution. The value displayed at each generation in each run is the average fitness over the entire population. It can be seen that the average fitness improves significantly over the 100 generations; however, it appears that convergence is not reached. To test this we also show how the fitness of the best individuals improves over time, in Figure~\ref{fig:BestFitness}. It can be seen that the best individuals reach fitness values above 3.5 and remain stable after approx. 50 generations. This confirms that the DE algorithm reaches convergence, and the apparent lack of convergence from Figure~\ref{fig:AvgFitnessAll} is due to averaging the values over all individuals in a population. 

In light of the above, we conclude that the behaviour of the fitness function is consistent with the typical behaviour of a DE algorithm, and we confirm that convergence is reached; therefore the optimisation method proposed is verified.

\subsection{Evaluation of the optimal system behaviour}
\begin{figure}
    \centering
    \includegraphics[width = .48\textwidth]{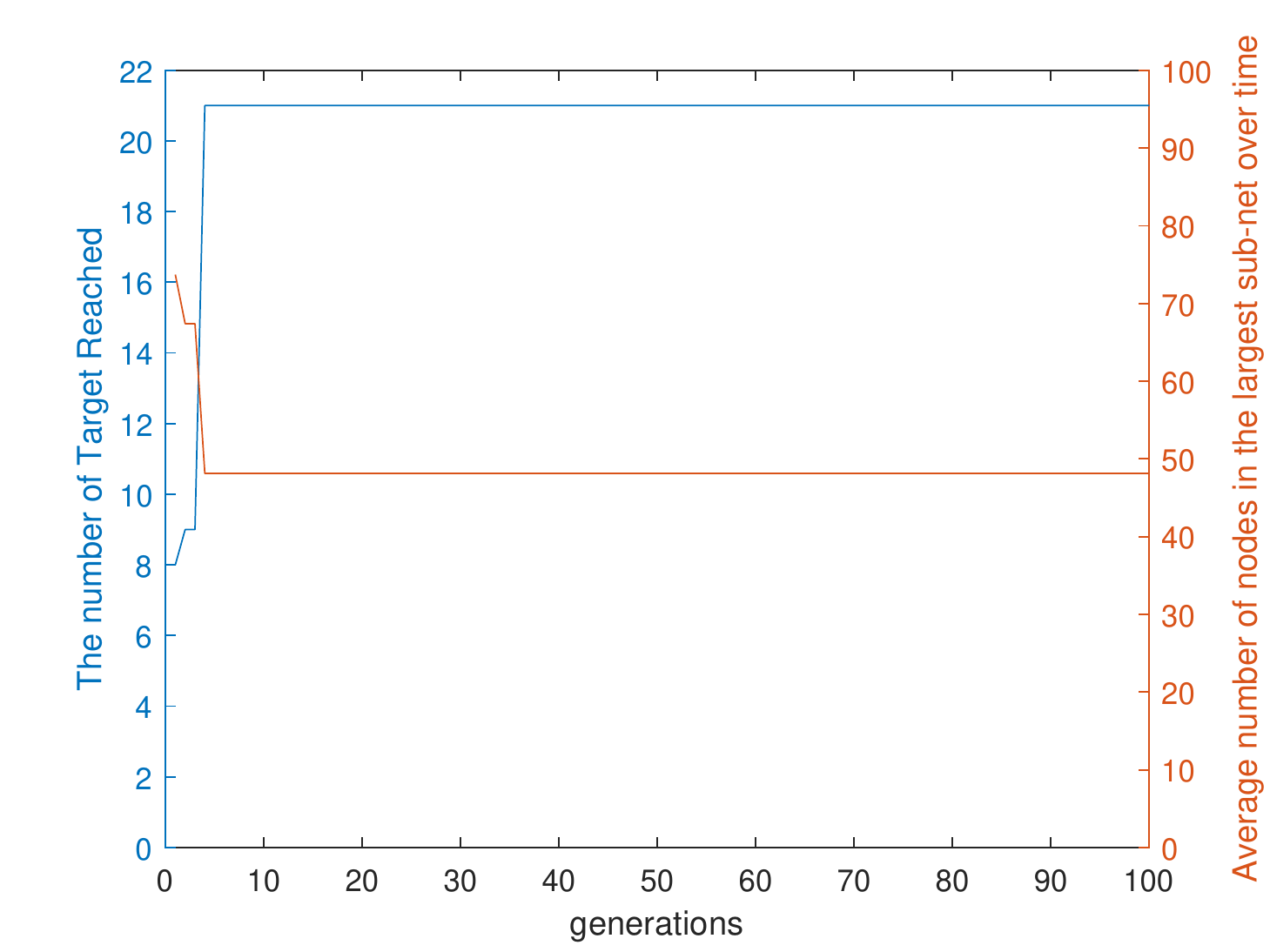}
    \caption{The individual with highest performance on reaching targets. Sample from one simulation run.}
    \label{fig:BestTarget}
\end{figure}

\begin{figure}
    \centering
    \includegraphics[width = .48\textwidth]{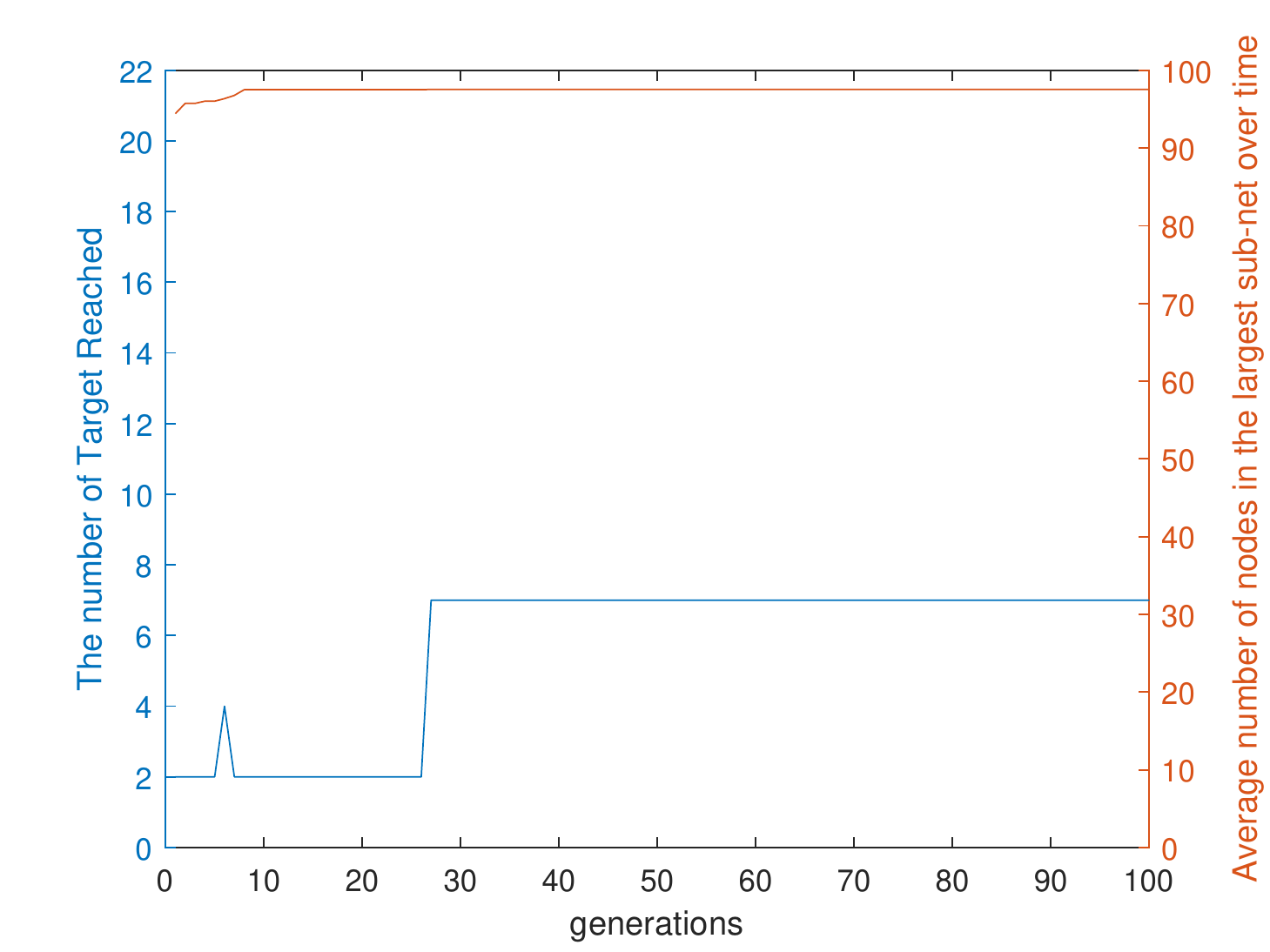}
    \caption{The individual with highest performance on connectivity. Sample from one simulation run.}
    \label{fig:BestNet}
\end{figure}

With the optimisation process verified, the next step is to discuss the performance of the optimal system, as resulting from the various values of the fitness function. First, we show two sets of results where we isolate the systems with highest value on each of the objectives (Figure~\ref{fig:BestTarget} and Figure~\ref{fig:BestNet}).

It becomes apparent that a trade-off exists between the two objectives. Figure~\ref{fig:BestTarget} shows the evolution of the system that reaches the highest number of targets (i.e. a total of 21 targets); however the connectivity represented through the largest giant component contains less than 50 ground agents. Figure~\ref{fig:BestNet} shows the opposite situation, when the system that maintains the highest level of connectivity (i.e. giant component contains over 97 ground agents), but can only reach 7 targets.

\begin{figure}
    \centering
    \includegraphics[width=.48\textwidth]{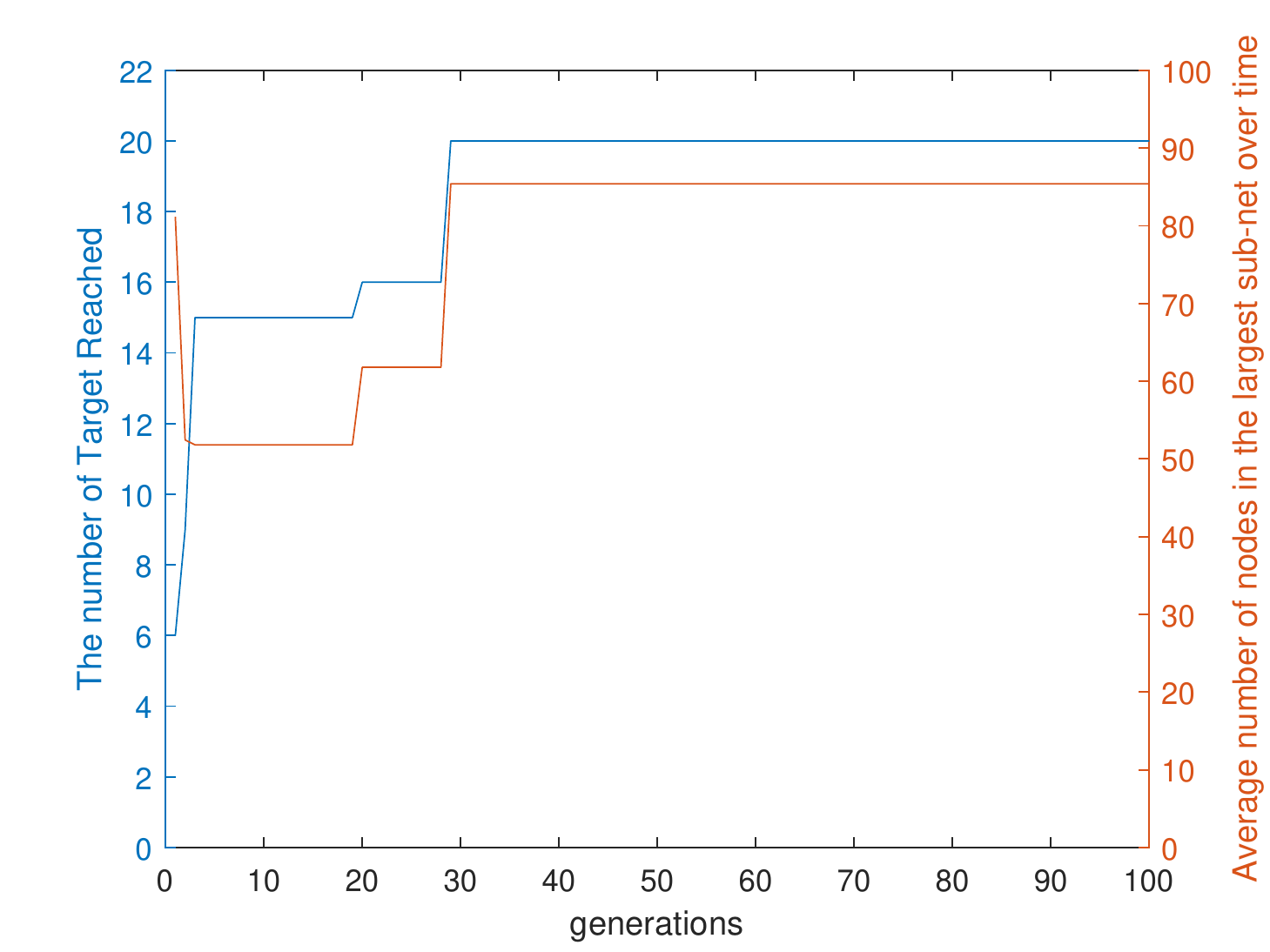}
    \caption{The best individual with overall performance. (The number of target reached is 20 and average percentage of nodes in the larges sub net is 85.40\%).}
    \label{fig:best_overall}
\end{figure}

We made no assumptions in the beginning that the objectives are in contradiction, however this trade-off can be explained through the way the two swarms operate in the field. It has been shown in~\cite{Leu2019cec} that a certain extend of grouping is necessary on the ground for the UAVs to be able to maintain high levels of ground connectivity. Since in this paper the swarm models we adopt are similar, we can conclude that a similar ground behaviour is necessary in our case too. However, in this paper we add target search, which involves a spreading of the ground agents in the field to be better distributed spatially and thus more likely to be in te proximity of a target.

In light of the above discussion we expect that the best performing system over both objectives, which corresponds to the highest value of the fitness function, shows an optimal trade-off between the two. Figure~\ref{fig:best_overall} confirms this, showing a system that is able to reach 20 targets while maintaining throughout the simulation on average a giant connected component of over 85 ground agents.

The visual results presented in Figures~\ref{fig:BestTarget},~\ref{fig:BestNet}~and~\ref{fig:best_overall} are further summarised in Table~\ref{tab:BestParameters} together with the values of the corresponding genes/parameters.

\begin{table*}[t]
\caption{The parameter values used by the best individuals and their performance}
    \centering
    \begin{tabular}{l|l|r r r} \hline 
\multirow{2}{*}{Agent Types}  & \multirow{2}{*}{Parameters} & \multicolumn{3}{ c } {Individuals}   \\ \cline{3-5}
 &  & Overall best & Best target tracking & Best network \\ \hline
Ground Agents & Alignment to air & 0.043 & 0.049 & 0.046 \\
 & cohesion to air & 0.000 & 0.000 & 0.001 \\\hline
Air agents & Alignment to ground & 0.627 & 0.656 & 0.730 \\
 & Alignment to air & 0.000 & 0.343 & 0.380 \\ 
 & Cohesion to ground & 0.566 & 0.384 & 0.407 \\
 & Cohesion to air & 0.055 & 0.000 & 0.046 \\
 & Separation  & 0.486 & 0.188 & 0.061 \\
 & Speed & 1.169 & 5.000 & 0.799 \\
 & Separation distance  & 290 & 114.605 & 290 \\ \hline
\multicolumn{2}{ c| }{number of targets detected} & 20 & 21 & 7\\ \hline
 \multicolumn{2}{ c| }{average nodes in the largest sub-net } & 85.4 & 48.14 & 97.54\\ \hline
 \end{tabular}
    \label{tab:BestParameters}
\end{table*}

%
%

\section{Conclusions}\label{Section:Conclusion}

In this paper we augment the existing literature on survivable networks by presenting a swarm teaming perspective on the narrow ``communication only'' problem typically addressed in this research domain. The paper considers a swarm of UAVs that not only facilitates connectivity between the agents of a ground swarm, but also participates directly in the ground task depending on that connectivity.

A target searching generic context is considered as test-bed, in which a swarm of ground agents and a swarm of UAVs cooperate so that the ground agents reach as many targets as possible in the field while also remaining connected as much as possible at all times. To optimise the system against both these objectives in the same time, an evolutionary computation approach was used, in the form of a differential evolution algorithm. Results obtained showed a good evolution of the fitness function used as part of the DE and a good performance of the evolved dual-swarm system.

The optimal system obtained through the bi-objective optimisation provided an optimal trade-off between target reaching and connectivity, thus validating the air-ground swarm teaming approach we proposed. While they are nevertheless bounded to the context chosen for investigation in this paper (which is one of the limitations that we acknowledge), the results are encouraging and open doors for numerous directions of investigation in the future.


\bibliographystyle{IEEEtran}
\bibliography{sscirefs}

\end{document}